\documentclass[a4paper]{article}

\usepackage{INTERSPEECH2021}

\usepackage{algorithm}
\usepackage{algorithmic}
\usepackage{hyperref}       % hyperlinks
\usepackage{url}   
\usepackage{comment}
\usepackage{color}
\usepackage{kotex}
\usepackage{adjustbox}

\usepackage{booktabs}       % professional-quality tables

\usepackage{multicol}
\usepackage{multirow}
\usepackage{amsmath}
\usepackage{mathtools}
\usepackage{amssymb}
\usepackage{comment}

\usepackage{caption}
\usepackage{subcaption}
\usepackage{xcolor}

\usepackage{balance}

\usepackage{url}
\renewcommand{\algorithmiccomment}[1]{\bgroup\hfill//~#1\egroup}
\title{PQK: Model Compression via Pruning, Quantization, and Knowledge Distillation}

\name{Jangho Kim$^{*,1,2}$\thanks{\scriptsize{${}^{*}$Author completed the research in part during an internship at Qualcomm Technologies, Inc. ${}^{\dagger}$Qualcomm AI Research is an initiative of Qualcomm Technologies, Inc.}}, Simyung Chang$^1$, Nojun Kwak$^{2}$}
\address{
  $^1$Qualcomm AI Research${}^{\dagger}$, Qualcomm Korea YH\\
  $^2$Seoul National University
  }
\email{kjh91@snu.ac.kr, simychan@qti.qualcomm.com, nojunk@snu.ac.kr}

\begin{document}

\maketitle
\begin{abstract}
As edge devices become prevalent, deploying Deep Neural Networks (DNN) on edge devices has become a critical issue. However, DNN requires a high computational resource which is rarely available for edge devices. To handle this, we propose a novel model compression method for the devices with limited computational resources, called \textit{PQK} consisting of pruning, quantization, and knowledge distillation (KD) processes. Unlike traditional pruning and KD, PQK makes use of unimportant weights pruned in the pruning process to make a teacher network for training a better student network without pre-training the teacher model. PQK has two phases. Phase 1 exploits iterative pruning and quantization-aware training to make a lightweight and power-efficient model.  In phase 2, we make a teacher network by adding unimportant weights unused in phase 1 to a pruned network. By using this teacher network, we train the pruned network as a student network. In doing so, we do not need a pre-trained teacher network for the KD framework because the teacher and the student networks coexist within the same network (See Fig. \ref{overallprocess}). We apply our method to the recognition model and verify the effectiveness of PQK on keyword spotting (KWS) and image recognition. 
\end{abstract}

\section{Introduction}

Nowadays, Deep Neural Networks (DNNs) have shown astonishing capabilities in various domains such as computer vision and signal processing. Although DNN shows remarkably high performance, it requires high computational cost and memory. Also, DNN models are spreading from personal computers or servers into edge devices. Deploying DNN on edge devices such as smartphones and IoT devices is still a challenge due to its computational resource constraint and restricted memory.

In recent years, model compression has been actively studied to deal with the above issues. In general, model compression can be categorized into three: pruning, quantization, and knowledge distillation. (1) \textit{Pruning} method prunes the unimportant weights or channels based on different criteria \cite{lin2020dynamic,han2015learning,He_2019_CVPR,kim2021prototypebased}. Pruning method can reduce model memory and the number of flops by eliminating unimportant weights or channels. (2) \textit{Quantization} method quantizes floating point values into discrete values to approximate them by a set of integers and scaling factors \cite{krishnamoorthi2018quantizing}. Quantization allows for more power-efficient operations and convolution computations at the expense of lower bitwidth representation. Recently, hardware accelerators such as NVIDIA’s Tensor Core and CIM (Compute-in-memory) devices have launched for 4-bit processing to improve the power efficiency \cite{pan2018multilevel,markidis2018nvidia}.
(3) \textit{Knowledge distillation} is a learning framework using teacher and student networks. Teacher network transfers its knowledge to student network to enhance the performance of student network. Feature maps \cite{NEURIPS2018_6d9cb7de,heo2019comprehensive,kim2021feature} and logits of a network \cite{hinton2015distilling,kim2019qkd} are widely used as knowledge. Model compression has actively been studied mainly on computer vision tasks. However, with the increase of various voice assistants such as Siri, Hey Google, and Alexa on IoT devices, model compression has also become an important research topic in speech processing \cite{Bai2020,Jose2020,Adya2020,Nguyen2020,chen2014small}. 

\begin{figure*}[t]
  \centering
   \includegraphics[width=0.8\linewidth]{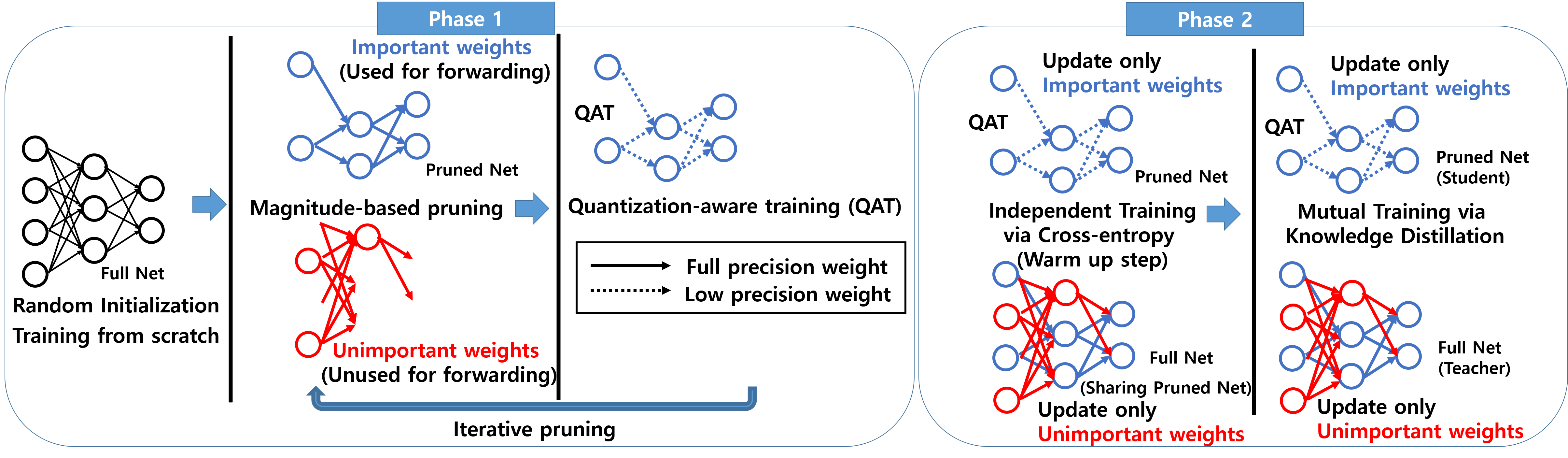}\\

  \caption{The overall process of PQK. Phase 1 trains the model from scratch with iterative pruning and quantization-aware training (QAT). The blue nodes and arrows corresponds to important weights used for the QAT, and the red ones are unimportant weights based on pruning method. The solid line and dotted line represent the full precision and $k$-bit quantized weights, respectively. At phase 2, we make a teacher network with a full network (Blue graph+Red graph). After some warm-up steps, we train the pruned net (student) and the full net (teacher) using KD framework. At teacher training, The blue graph from the student is fixed and shared in the full net, and the the red graph is only updated. However, the blue graph is shared at the forwarding of the full net.}
    \label{overallprocess}
\end{figure*}

In this work, we aim to design the PQK to leverage pruning, quantization, and knowledge distillation by considering each method's characteristics. In contrast with traditional pruning and knowledge distillation, we use unimportant weights considered in the pruning process to make a teacher network, so PQK does not need a pre-trained teacher model. We propose PQK to compress the keyword spotting (KWS) recognition model. PQK can also be used for the image recognition model because the design of PQK is focused on the training framework regardless of datatype and model, which has high applicability. PQK consists of two phases. In the first phase, we train the model from scratch using both iterative pruning and quantization-aware training (QAT). We prune the model and quantize the pruned model with a learnable step-size for QAT. This phase focus on finding a pruned model from scratch together with quantizing the model. In phase 2, we make a teacher network called full net shown in Fig. \ref{overallprocess} by combining the pruned net and the unused weights considered unimportant in phase 1. Then, we train the pruned network as a student network. This phase improves the performance of the pruned net (student) by knowledge distillation with the full net (teacher). The details of PQK are explained in Sec.~\ref{sec:method} and Fig. \ref{overallprocess}. 

\section{Proposed Method}
\label{sec:method}
In this section, we first explain the pruning and quantization method used in PQK and then describe the proposed PQK.
\subsection{Preliminaries}
\noindent \textbf{Pruning } Iterative pruning \cite{lin2020dynamic} is widely used in machine learning because it generally outperforms the one-shot pruning method \cite{han2015learning}. One-shot pruning just prunes the model once with specific sparsity after model training. Then, it finetunes the model to improve the performance of the pruned model. On the other hand, iterative pruning gradually prunes the model while training and the final model contains unpruned important weights. In this work, we use iterative pruning and adopt the gradually increasing pruning ratio scheme based on the current epoch ($c$) \cite{lin2020dynamic}:    
\begin{equation} 
p_c = p_t + (p_i-p_t)(1-\frac{c-c_0}{n})^3.
\label{eq:prunning_ratio}
\end{equation} 
We increase the pruning ratio from an initial ratio ($p_i=0$) to a target pruning ratio $p_t$ over the training epoch ($n$). $p_c$ represents the current pruning ratio per epoch $c \in \{c_0, ..., c_0+n\}$, where $c_0$ means an initial epoch ($c_0=0$). 

\noindent \textbf{Quantization } 
After pruning step, we train the model with quantization-aware training (QAT) using important weights. We choose the uniform symmetric quantization method and the per-layer quantization scheme considering hardware friendliness \cite{krishnamoorthi2018quantizing,NEURIPS2020_eb1e7832}. Consider the range of model weight [$min_w$,$max_w$]. The weight $w$ is quantized to an integer value $\hat{w}$ with the range of [$-2^{k-1}+1$,$2^{k-1}-1$] according to $k$-bit. Quantization and dequantization for the weight are defined with the learnable stepsize $S_w$. The overall quantization process is as follows:
\begin{equation}
 \hat{w}=Clip(\Bigl\lfloor{\frac{w}{S_w}}\Bigr\rceil,-2^{k-1}+1, 2^{k-1}-1),
 \label{Eq:k_bit}
\end{equation} 
where $\lfloor \cdot \rceil$ is the round operation and 
%\begin{align*}

$
Clip(w, a, b) = \begin{cases}
    b & \text{if } \quad w > b \\
    a & \text{if } \quad w < a \\
    w & \text{otherwise}.
    \end{cases}
$

Dequantization step just brings the quantized value back to the original range by multiplying the step-size:
\begin{equation}
 \bar{w}=\hat{w} \times S_w.
\end{equation} 
These quantization and dequantization processes are non-differentiable, so we utilize a straight-through estimator (STE) \cite{bengio2013estimating} for backpropagation. STE approximates the gradient $\frac{d\bar{w}}{d{w}}$ by 1. Therefore, we can approximate gradients $\mathcal{L}$, $\frac{d\mathcal{L}}{d{w}}$, with $\frac{d\mathcal{L}}{d\bar{w}}$. 
\begin{equation} 
\frac{d\mathcal{L}}{d{w}} = \frac{d\mathcal{L}}{d\bar{w}}\frac{d\bar{w}}{d{w}} \approx \frac{d\mathcal{L}}{d\bar{w}}.
\end{equation}

\begin{algorithm}[t]
 \algsetup{linenosize=\scriptsize}
 \scriptsize
\centering
\renewcommand{\algorithmicrequire}{\textbf{Input:}}
\renewcommand{\algorithmicensure}{\textbf{Output:}}
\renewcommand{\algorithmicprint}{\textbf{break}}
\caption{\text{ PQK}}
\label{alg:pqk}

\begin{algorithmic}[1]
\REQUIRE Untrained model $\mathbf{W}$; \\ Number of epochs for each phase $P_1$, $P_2$;
\\ Number of iterations for mask update $p_u$ and number of epochs for warm up stage $s$; \\ pruning mask $\mathcal{M}$, Step-size $S_w$
\ENSURE Trained model (Full Net) $\mathbf{W}$ and pruned and quantized model (Pruned Net) $\mathbf{\bar{W}}\odot\mathcal{M}$
\STATE \textbf{Phase 1: Pruning and Quantization} 
\FOR{ Epoch = 1 ,..., $P_1$}
\STATE compute sparsity $p_c$ (\ref{eq:prunning_ratio})
\FOR{Iter = 1 ,..., N}
\IF{Iter $\%$ $p_u$==0 } 
\STATE compute mask $\mathcal{M}$ with $p_c$ and magnitude pruning \COMMENT{Update mask every $p_u$ iteration}
\ENDIF
\STATE Update $S_w$, $\mathbf{W}$ (\ref{Eq:phase1})  by minimizing $\mathcal{L}^{S}_{ce}$ (\ref{eq:Lce})
 %using backpropagation
\ENDFOR
\ENDFOR
\STATE \textbf{Phase 2: Knowledge Distillation}
\STATE init $\alpha=1, \beta=0$. $S_w$ and $\mathcal{M}$ are fixed  \COMMENT{Warm up stage only uses cross-entropy}
\FOR{ Epoch = 1 ,..., $P_2$} 
\IF{$s$ $<$ Epoch} 
\STATE set $\alpha,\beta$ \COMMENT{KD training after warm up stage}
\ENDIF
\FOR{Iter = 1 ,..., N}
\STATE Update $\mathbf{W}$ (\ref{Eq:phase2_student},\ref{Eq:phase2_teacher})  by minimizing $\mathcal{L}^{S}_{KD}$ (\ref{eq:LKDS}) and $\mathcal{L}^{T}_{KD}$ (\ref{eq:LKDT})
\ENDFOR
\ENDFOR
\end{algorithmic}
\end{algorithm}

\vspace{-6.5mm}
\subsection{PQK}
\noindent \textbf{Notations } %We clarify the notation of this work. 
Consider a Convolutional Neural Network (CNN) with $L$ layers\footnote{In our notation, a layer contains the corresponding weights.} as an example. Then, we can represent the weights of the CNN model as $\{w_l : 1 \leq l \leq L\}$. To represent the pruned model with a binary matrix, we use $\{\mathcal{M}_l : 1 \leq l \leq L\}$. Each $\mathcal{M}_l$, is a binary matrix indicating whether they are pruned or not. Set $\mathcal{I}_l$ is all indices of $w_l$ at the $l$-th layer. $\mathcal{I}_{\mathcal{M}_l}$ and $\mathcal{I}_{\sim\mathcal{M}_l}$ indicate indices of the important weights (Blue graph in Fig. \ref{overallprocess}) and unimportant weights (Red graph in Fig. \ref{overallprocess}) at the $l$-th layer, respectively ($\mathcal{I}_l= \mathcal{I}_{\mathcal{M}_l} \cup \mathcal{I}_{\sim\mathcal{M}_l} $).

Assuming that we handle a recognition task with $m$ classes, the logit vector of a model is defined as $\mathbf{z}^t$, where $t \in \{S, T\}$ is the type of the network, i.e. either the student or the teacher. The network can have different paths depending on the target bit-width $k$. We can consider the pruned network as a student network whose path is determined by the masks $\{\mathcal{M}_l\}_{l=1}^L$ and the full network as a teacher network which utilizes all weights (important + unimportant weights). At phase 2, we make a teacher network using both unimportant weights and important weights. Then, we make a soft probability distribution with temperature $\mathcal{T}$ as:
\begin{equation} 
\sigma_a(\mathbf{z}^t;\mathcal{T}) = \frac{ e^{z^t_a/\mathcal{T}}}{\sum_{b}^me^{z^t_b/\mathcal{T}}},\:\: t \in \{S,T\}.
\end{equation} 
Here, $\mathbf{z}^S$ is the logit forwarded by blue graph and $\mathbf{z}^T$ is the logit from blue+red graph, shown in Fig. \ref{overallprocess}. Based on this notation, we can define the cross-entropy loss as below:
\begin{equation} 
\mathcal{L}_{ce}^t = -\sum_{a=1}^{m}y_a\log(\sigma_a(\mathbf{z}^t;1)), \:\: t \in \{S,T\}
\label{eq:Lce}
\end{equation}
where $y$ is a ground truth and the subscript $a$ denotes the $a$-th element of the corresponding vector.

\noindent \textbf{Phase 1 } Generally, phase 1 has the same number of epochs compared to conventional training and trains model from scratch. At phase 1, PQK combines iterative pruning and the quantization-aware training. First, PQK prunes the model at some epochs based on the pruning ratio in Eq. (\ref{eq:prunning_ratio}) by magnitude-based unstructured pruning \cite{han2015learning}. It calculates the pruning mask $\mathcal{M}$ which acts as gate functions. Note that, PQK update the pruning mask every $p_{u}$-th iteration similar to \cite{lin2020dynamic}. Then, QAT is performed with important weights using trainable step-size $S_w$. By using STE and the chain rule, the update rule at the $l$-th layer becomes  
\begin{equation}
    {w}^{(i,j)}_{l} \leftarrow {w}^{(i,j)}_{l} - \eta 
      \frac{\partial \mathcal{L}_{ce}^S}{\partial \bar{w}^{(i,j)}_{l} \mathcal{M}^{(i,j)}_{l}}, \:\: \forall(i,j) \in \mathcal{I}_l
\label{Eq:phase1}
\end{equation}
where, $(i,j)$ is a index of weight matrix. Note that, PQK also updates $S_w$ with $\mathcal{L}_{ce}^S$. As depicted in Fig \ref{overallprocess}, $\mathcal{L}_{ce}^S$ is calculated by forwarding only important weights. 

\noindent \textbf{Phase 2 } At phase 2, PQK trains the full network and pruned network with additional epochs. Commonly, to leverage knowledge distillation, a pre-trained teacher model is needed. Unlike traditional KD, PQK makes a teacher model with unimportant weights which means unused weights at phase 1 (See Fig \ref{overallprocess}). Note that there is no pre-trained teacher network because the teacher and student network are in the same network (full net).

We can compute the Kullback–Leibler divergence (KL) between student and teacher network. 
\begin{equation} 
\textit{KL}(\mathbf{z}^T||\mathbf{z}^S; \mathcal{T}) = \sum_{a=1}^{m}\sigma_a(\mathbf{z}^T;\mathcal{T})\log(\frac{\sigma_a(\mathbf{z}^T;\mathcal{T})}{\sigma_a(\mathbf{z}^S;\mathcal{T})})
\label{eq:KLloss}
\end{equation} 
 Then, we update each network with cross entropy and KL loss as below:
\begin{equation} 
\mathcal{L}^{S}_{KD} = \alpha\mathcal{L}_{ce}^S + \beta(\mathcal{T}^2\ast{\textit{KL}}(\mathbf{z}^T||\mathbf{z}^S; \mathcal{T}))
\label{eq:LKDS}
\end{equation} 
\begin{equation} 
\mathcal{L}^{T}_{KD} = \alpha\mathcal{L}_{ce}^T + \beta(\mathcal{T}^2\ast{\textit{KL}}(\mathbf{z}^S||\mathbf{z}^T; \mathcal{T}))
\label{eq:LKDT}
\end{equation} 
$\mathcal{L}^{S}_{KD}$ and $\mathcal{L}^{T}_{KD}$ are the KD loss of pruned net (student) and full net (teacher), respectively. $\alpha$ and $\beta$ are hyper-parameters for balancing between $\textit{KL}$ and cross-entropy losses. $\mathcal{T}^2$ is multiplied to the KL loss because the gradient with respect to the logit decrease as much as $1/\mathcal{T}^2$. The update rules at the $l$-th layer becomes
\begin{equation}
    {w}^{(i,j)}_{l} \leftarrow {w}^{(i,j)}_{l} - \eta 
      \frac{\partial \mathcal{L}_{KD}^S}{\partial \bar{w}^{(i,j)}_{l} \mathcal{M}^{(i,j)}_{l}},\:\: \forall(i,j) \in \mathcal{I}_{\mathcal{M}_l}
      \label{Eq:phase2_student}
\end{equation}
\begin{equation}
    {w}^{(i,j)}_{l} \leftarrow {w}^{(i,j)}_{l} - \eta 
      \frac{\partial \mathcal{L}_{KD}^T}{\partial {w}^{(i,j)}_{l}},\:\: \forall(i,j) \in \mathcal{I}_{\sim\mathcal{M}_l}
      \label{Eq:phase2_teacher}
\end{equation}
Note that, with respect to the pruned network, based on Eq. (\ref{Eq:phase2_student}) keeping the same bitwidth of phase 1, phase 2 updates only important weights unlike phase 1 (Eq. (\ref{Eq:phase1})) updating all weights. Analogous to the pruned network, in terms of the full network, phase 2 updates only unimportant weights (Eq. (\ref{Eq:phase2_teacher})). At the forwarding path of the full network, pruned network is shared and the full network does not use QAT. Also, we fix $S_w$ at phase 2 for a stable training.

At first few epochs, we set the hyper-parameters as $\alpha=1,\beta=0$ meaning that both pruned and full nets are trained by cross-entropy only because initial unimportant weights are not trained well at phase 1. Thus, it needs a warm up stage. The overall process of PQK is depicted in Fig \ref{overallprocess} and Algorithm \ref{alg:pqk}.

\section{Experiments}
We verify the proposed PQK on a keyword spotting task. Also, we conduct additional experiments on an image recognition task to show the applicability and generality of the proposed PQK. We set the target pruning ratio $p_t$ (Eq. (\ref{eq:prunning_ratio})) as 0.9 that means we only use 10\% parameters of the baseline model. For ResNet-8 \cite{tang2018deep}, we also conduct various target pruning ratio ($p_t\in\{0.9,0.7,0.5\}$). We quantize the model by 8-bit and 4-bit ($k\in\{8,4\}$, Eq. (\ref{Eq:k_bit})) compared to the 32-bit baseline model. Although we have one network in PQK, at phase 2, we have twice forwarding for pruned net and full net, so they need different batch statistics in phase 2. Therefore, we use different batchnorm parameters for each net in phase 2. 

\subsection{Experimental Setup}
In all experiments, we use pytorch framework and set the same hyper-parameters. We update the pruning mask per 32 iterations ($p_u$) at phase 1. After the warm up stage in phase 2, we set $\mathcal{T}=2, \alpha=0,5, \beta=0.5$. We did not conduct a grid search for finding hyper-parameters but choose them based on recommendations from related works \cite{tucker2016model,kim2019qkd,lin2020dynamic}. For the learning rate of learnable step-size $S_w$, we multiply $10^{-4}$ to the initial learning rate of model parameters because of its sensitivity.     

\noindent \textbf{Keyword Spotting }
We use Google’s Speech Commands Dataset v1 \cite{Pete}, choosing ResNet-8 and ResNet-8-narrow \cite{tang2018deep} as baselines using the official code in pytorch\footnote{\url{https://github.com/castorini/honk}}. At phase 1, we follow overall training details from \cite{tang2018deep}. At phase 2, we run 9 epochs for additional training. We start learning rate of 0.1 and decay it at 1000 and 2000 iteration by multiplying 0.1. We set the warm up iteration ($s$) as 1500.

\noindent \textbf{Image Recognition }
We use CIFAR100 Dataset \cite{krizhevsky2009learning} and choose ResNet-32 \cite{he2016deep} for the baseline. We follow the training details same as \cite{he2016deep}. At phase 2, we use 60 additional epochs. We train the model with an initial learning rate of 0.2 and decay it at 20,40 epochs with a factor of 0.1. We start KD after 30 epochs ($s = 30$).

\begin{table*}[t]

\centering
  \caption{ 
  Test accuracy with various setting on speech and image dataset.
  }
\label{main_result}
\begin{adjustbox}{width=0.75\linewidth}
   \begin{tabular}{  l c c  |c | c | c | c| c }
         \toprule
        & &  &  CIFAR100 & \multicolumn{4}{c}{Google’s Speech Commands Dataset}  \\
        \midrule
          &    & & \multicolumn{3}{c|}{$p_t=0.9$}  &  $p_t=0.7$ & $p_t=0.5$   \\  
         \multirow{3}{*}{Method}& \multirow{3}{*}{Bitwidth} & \multirow{3}{*}{Pruning ratio} & \multicolumn{1}{c|}{ ResNet-32} & \multicolumn{1}{c|}{ ResNet-8-narrow} & \multicolumn{1}{c}{ ResNet-8} & \multicolumn{1}{|c|}{ ResNet-8} & \multicolumn{1}{|c}{ ResNet-8} \\
           
                &  & & {Accuracy (\%)}   & {Accuracy (\%)} & {Accuracy (\%)} & {Accuracy (\%)} & {Accuracy (\%)}   \\
         \midrule
   
            Vanilla  & 32   & 0  & 69.7 & 91.4   & 94.3 & 94.3 & 94.3     \\
                      Phase1-$P$  &  8 & $p_t$   & 67.4  & 81.7   & 92.6& 94.3&  94.3    \\
                      
                      Phase2-$P$  &  8 & $p_t$  & 69.8   & 86.4   & 94.0   & 94.6 &  94.7 \\
                      
                      Phase2-$F$  &  32 ($P=8$) & 0   & 71.1   & 90.1   & 94.4 & 94.8 &  94.9  \\
                      Phase1-$P$  &  4 & $p_t$   & 66.2    & 74.1  & 91.7  & 94.3 & 94.1 \\
                      
                     Phase2-$P$  &  4 & $p_t$   & 67.7 & 83.1   & 93.6 & 94.1 &  94.6   \\
                      
                      Phase2-$F$  &  32 ($P=4$) & 0   & 69.8  & 85.1   & 93.4    & 93.2 & 93.7  \\
 \bottomrule

   \end{tabular}   
%\end{adjustbox}
\end{adjustbox}
\end{table*}

\subsection{Experimental Results}
In this section, we show the results of PQK with various methods, bitwidths, and pruning ratios. %used in model parameters. 
We will refer to the baseline network which is trained with cross-entropy as vanilla in all experiments. There are two forwarding paths in the output of phase 2. The first one, pruned net (student), uses $(1-p_t)\times100$\% and quantized parameters. The other one, full net (teacher), uses the whole parameters same as the vanilla. At every table, we refer to our method at the end of each phase and network type $\in \{P,F\}$, where $P$ and $F$ represent pruned net and full net, respectively. Full net contains and shares the pruned net so the bitwidth of full net is 32-bit containing the pruned net trained with $k$ bitwidth QAT. For example, at the 4-th row in Table \ref{main_result}, phase2-$F$ and 32 ($P=8$) means full net of phase 2 with 32-bitwidth, sharing pruned net trained with 8-bitwidth QAT.    

\noindent \textbf{Keyword Spotting }
As shown in Table ($p_t=0.9$), the performance of phase1-$P$ decreases compared to vanilla. This is because, in this phase, we pruned unimportant 90\% weights of full net and quantize important 10\% weights from 32-bit to 8-bit or 4-bit using iterative pruning and QAT. A compact model, ResNet-8-narrow using fewer channels than ResNet-8, is more sensitive to the model compression. It degrades 9.7\% and 17.3\% at 8-bit and 4-bit in phase1-$P$. Such severe performance degradation of the compact model with quantization is also reported in other researches \cite{hubara2016binarized,kim2019qkd}. At phase 2, by training unimportant 90\% weights of the full net, it becomes a teacher to improve the performance of the pruned net. Surprisingly, there is a large performance gap between phase 1 and 2 in the pruned model of ResNet-8-narrow compared to that of ResNet-8. Table \ref{main_result} shows that PQK is more effective at the compact model in terms of recovering the decreased performance at phase 1, where performance enhancements are 4.7\% and 9\% at 8- and 4-bit in ResNet-8-narrow. Concerning bitwidth and accuracy, 8-bit consistently outperforms 4-bit because of its high representative power from more bitwidth. In ResNet-8, Regardless of bitwidth, as pruning ratio decreases, the performance of pruned net increases. These numbers show the usage of model parameters is important to the performance. In ResNet-8 with 4-bit, although pruned net performs well, accuracies of phase2-F are lower than those of phase2-P. ResNet-8 with 8-bit shows the opposite trend, meaning that combining 32-bit and 4-bit trained model is more unstable than combining 32-bit and 8-bit trained model.

\noindent \textbf{Image Recognition }
In this experiment, we can show the applicability of PQK. The design of PQK is not dependent on dataset and model architecture because 
PQK prunes and quantizes the model regardless of dataset and model architecture. Image recognition task has a similar tendency with KWS task. Interestingly, the performance of phase2-$F$ containing 8-bit pruned net outperforms vanilla by 1.4\%. At phase 2, the teacher network is also trained with KD using the student network (Eq. (\ref{eq:LKDT})). In doing so, the full net can outperform the vanilla. 

\begin{figure}[t]
  \centering
   \includegraphics[width=0.75\linewidth]{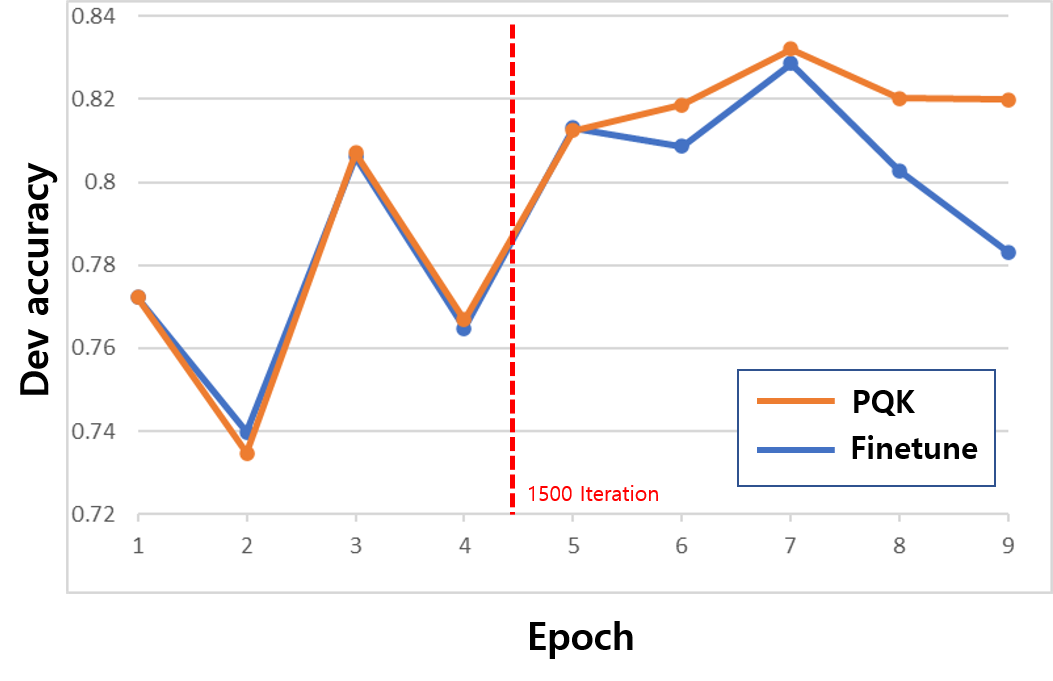}\\

  \caption{Dev accuracy of ResNet-8-narrow on google's speech command dataset at every epoch :Orange line represents PQK (Phase2-P) and blue line shows the finetune (lr=0.1). Red dot line means the end of warm up iteration.   }
    \label{ablation_study_figure}
%\vspace{-2mm}
\end{figure}

\noindent \textbf{Ablation study }
To show the effectiveness of phase 2 in PQK, we conduct an ablation study in Table \ref{ablation_study_table}. At phase 2, we have additional epochs for the boosting performance of the pruned net. We make various baselines using the same training budget as phase 2. Finetune in Table \ref{ablation_study_table} represents the performance of finetuning 4-bit ResNet-8-narrow model from phase1-$P$  (Table \ref{main_result}) with additional training using various learning rates. We use the same experiment setting with phase2-$P$ using 9 epochs and decay learning rate at 1000 and 2000 iteration. The only difference is the existence of KD with full net. In finetuning methods, the high learning rate is more efficient than the lower one, where 0.1 shows the best performance along with various learning rates. However, phase 2 using KD framework utilizing unused weights in phase 1 outperforms the best finetuning method by 3.9\%. We also plot the validation accuracy of finetuning and phase-2 of PQK per every epoch in Fig. \ref{ablation_study_figure}. In this figure, orange and blue line represent the validation accuracy of phase2-$P$ and finetune (lr=0.1) in Table \ref{ablation_study_table}. During the warm up step, two methods show very similar trends because they are trained with only cross-entropy. After warm up step, the performance gap between them increases because mutual KD training helps to enhance the performance of both pruned and full net.

\begin{table}[t]

\centering
  \caption{%ResNet-56 on CIFAR-10/100 dataset. %C-10 and C-100 denote CIFAR-10 and CIFAR-100, respectively. 
  Ablation study on PQK: comparing PQK (phase2-$P$) with finetuning using same training budget on various learning rate, where all methods start from phase1-$P$. 
  }
\label{ablation_study_table}
\begin{adjustbox}{width=0.8\linewidth}
  \begin{tabular}{  l c c c}
         \toprule
          \multicolumn{4}{c}{ Google’s Speech Commands Dataset} \\

          \multicolumn{4}{c}{ ResNet-8-narrow} \\
          \midrule
              {Method}  & Bitwidth & {Accuracy (\%)} & {Pruning ratio}  \\
         \midrule
  Phase1-$P$ & 4   & 74.1   & 0.9     \\
            Finetune (lr=0.1) & 4   & 79.2   & 0.9     \\
            Finetune (lr=0.01)& 4   & 78.0   & 0.9     \\
            Finetune (lr=0.001)& 4   & 73.8   & 0.9     \\
                      Phase2-$P$  &  4 & 83.1   & 0.9     \\
                                              
        \bottomrule

  \end{tabular}   
%\end{adjustbox}
\end{adjustbox}
\end{table}

\section{Conclusions}
We propose a novel model compression framework to cope with the limited computational resource. This is a new way of model compression by leveraging pruning, quantization, and knowledge distillation. In phase 1, we are combining pruning and quantization to make a lightweight and power-efficient model. Then, in phase 2, we boost the performance of a efficient model by KD. We verify the efficiency of PQK on KWS and image recognition tasks. 

% \noindent\textbf{Acknowledgement} Nojun Kwak was supported by the National Research Foundation of Korea (NRF) grant funded by the Korea government (2021R1A2C3006659).
\clearpage

\bibliographystyle{IEEEtran}
\balance
\bibliography{mybib}

\end{document}